\documentclass[11pt]{article}
\usepackage{acl2016}
\usepackage{url}
\usepackage{times}
\usepackage{latexsym}
\aclfinalcopy 

\usepackage[utf8]{inputenc}
\usepackage[T1]{fontenc}

\usepackage{enumitem}
\usepackage{amsmath}
\usepackage{amssymb}
\usepackage{stmaryrd}
\usepackage{multirow}
\usepackage{tikz-dependency}

\usepackage{siunitx}
\usepackage{pgfplots}
\pgfplotsset{compat=newest}

\renewcommand\footnotemark{}
\title{Linguistic Input Features Improve Neural Machine Translation}

\author{
Rico Sennrich \and Barry Haddow\\
School of Informatics, University of Edinburgh\\
{\tt rico.sennrich@ed.ac.uk}, {\tt bhaddow@inf.ed.ac.uk}
}

\date{}

\begin{document}
\maketitle
\begin{abstract}

Neural machine translation has recently achieved impressive results, while using little in the way of external linguistic information.
In this paper we show that the strong learning capability of neural MT models does not make linguistic features redundant; they can be easily incorporated to provide further improvements in performance. 
We generalize the embedding layer of the encoder in the attentional encoder--decoder architecture to support the inclusion of arbitrary features, in addition to the baseline word feature.
We add morphological features, part-of-speech tags, and syntactic dependency labels as input features to English$\leftrightarrow$German and English$\to$Romanian neural machine translation systems.
In experiments on WMT16 training and test sets, we find that linguistic input features improve model quality according to three metrics: perplexity, {\sc Bleu} and {\sc chrF3}. 
An open-source implementation of our neural MT system is available\footnote{\url{https://github.com/rsennrich/nematus}}, as are sample files and configurations\footnote{\url{https://github.com/rsennrich/wmt16-scripts}}.
\end{abstract}

\section{Introduction}

Neural machine translation has recently achieved impressive results \cite{DBLP:journals/corr/BahdanauCB14,jean15b},
while learning from raw, sentence-aligned parallel text and using little in the way of external linguistic information.\footnote{Linguistic tools are most commonly used in preprocessing, e.g.\ for Turkish segmentation \cite{DBLP:journals/corr/GulcehreFXCBLBS15}.}
However, we hypothesize that various levels of linguistic annotation can be valuable for neural machine translation.
Lemmatisation can reduce data sparseness, and allow inflectional variants of the same word to explicitly share a representation in the model.
Other types of annotation, such as parts-of-speech (POS) or syntactic dependency labels, can help in disambiguation.
In this paper we investigate whether linguistic information is beneficial to neural translation models, or whether their strong learning capability makes explicit linguistic features redundant.

Let us motivate the use of linguistic features using examples of actual translation errors by neural MT systems.
In translation out of English, one problem is that the same surface word form may be shared between several word types,
due to homonymy or word formation processes such as conversion.
For instance, \emph{close} can be a verb, adjective, or noun, and these different meanings often have distinct translations into other languages.
Consider the following English$\to$German example:

\begin{enumerate}[resume] 
\item \textit{We thought a win like this might be close.} \label{en-de-example1}
\item \textit{Wir dachten, dass ein solcher Sieg nah sein könnte.} \label{en-de-example2}
\item *\textit{Wir dachten, ein Sieg wie dieser könnte schließen.} \label{en-de-example3}
\end{enumerate}

For the English source sentence in Example \ref{en-de-example1} (our translation in Example \ref{en-de-example2}), a neural MT system (our baseline system from Section \ref{sec:eval}) mistranslates \emph{close} as a verb, and produces the German verb \emph{schließen} (Example \ref{en-de-example3}), even though \emph{close} is an adjective in this sentence, which has the German translation \emph{nah}.
Intuitively, part-of-speech annotation of the English input could disambiguate between verb, noun, and adjective meanings of \emph{close}.

As a second example, consider the following German$\to$English example:

\begin{enumerate}[resume] 
\item \textit{Gefährlich ist die Route aber dennoch .}\\
dangerous is the route but still . \label{de-en-example1}
\item \textit{However the route is dangerous .} \label{de-en-example2}
\item *\textit{Dangerous is the route , however .} \label{de-en-example3}
\end{enumerate}

German main clauses have a verb-second (V2) word order, whereas English word order is generally SVO.
The German sentence (Example \ref{de-en-example1}; English reference in Example \ref{de-en-example2}) topicalizes the predicate \emph{gefährlich} 'dangerous', putting the subject \emph{die Route} 'the route' after the verb.
Our baseline system (Example \ref{de-en-example3}) retains the original word order, which is highly unusual in English, especially for prose in the news domain.
A syntactic annotation of the source sentence could support the attentional encoder-decoder in learning which words in the German source to attend (and translate) first.

We will investigate the usefulness of linguistic features for the language pair German$\leftrightarrow$English, considering the following linguistic features:

\begin{itemize}

\item lemmas
\item subword tags (see Section \ref{subword-sec})
\item morphological features
\item POS tags
\item dependency labels
\end{itemize}

The inclusion of lemmas is motivated by the hope for a better generalization over inflectional variants of the same word form.
The other linguistic features are motivated by disambiguation, as discussed in our introductory examples.

\section{Neural Machine Translation}

We follow the neural machine translation architecture by \newcite{DBLP:journals/corr/BahdanauCB14}, which we will briefly summarize here.

The neural machine translation system is implemented as an attentional encoder-decoder network with recurrent neural networks.

The encoder is a bidirectional neural network with gated recurrent units \cite{cho-EtAl:2014:EMNLP2014} that reads an input sequence $x=(x_1,...,x_m)$ and calculates a forward sequence of hidden states $(\overrightarrow{h}_1,...,\overrightarrow{h}_m)$,
and a backward sequence $(\overleftarrow{h}_1,...,\overleftarrow{h}_m)$.
The hidden states $\overrightarrow{h}_j$ and $\overleftarrow{h}_j$ are concatenated to obtain the annotation vector $h_j$.

The decoder is a recurrent neural network that predicts a target sequence $y=(y_1,...,y_n)$.
Each word $y_i$ is predicted based on a recurrent hidden state $s_i$, the previously predicted word $y_{i-1}$, and a context vector $c_i$.
$c_i$ is computed as a weighted sum of the annotations $h_j$.
The weight of each annotation $h_j$ is computed through an \emph{alignment model} $\alpha_{ij}$, which models the probability that $y_i$ is aligned to $x_j$.
The alignment model is a single-layer feedforward neural network that is learned jointly with the rest of the network through backpropagation.

A detailed description can be found in \cite{DBLP:journals/corr/BahdanauCB14}, although our implementation is based on a slightly modified form of this architecture, released for the dl4mt 
tutorial\footnote{\url{https://github.com/nyu-dl/dl4mt-tutorial}}. 
Training is performed on a parallel corpus with stochastic gradient descent.
For translation, a beam search with small beam size is employed.

\subsection{Adding Input Features}

Our main innovation over the standard encoder-decoder architecture is that we represent the encoder input as a combination of features \cite{alexandrescu-kirchhoff:2006:HLT-NAACL06-Short}.

We here show the equation for the forward states of the encoder (for the simple RNN case; consider \cite{DBLP:journals/corr/BahdanauCB14} for GRU):

\begin{equation}
\overrightarrow{h}_j = \text{tanh}(\overrightarrow{W}Ex_j + \overrightarrow{U}\overrightarrow{h}_{j-1})
\end{equation}

where $E \in \mathbb{R}^{m\times K_x}$ is a word embedding matrix, $\overrightarrow{W} \in \mathbb{R}^{n\times m}$, $\overrightarrow{U} \in \mathbb{R}^{n\times n}$ are weight matrices,
with $m$ and $n$ being the word embedding size and number of hidden units, respectively, and $K_x$ being the vocabulary size of the source language.

We generalize this to an arbitrary number of features $|F|$:

\begin{equation}
\overrightarrow{h}_j = \text{tanh}(\overrightarrow{W}(\bigparallel_{k=1}^{|F|}{E_k x_{jk}}) + \overrightarrow{U}\overrightarrow{h}_{j-1})
\end{equation}

where $\parallel$ is the vector concatenation,
$E_k \in \mathbb{R}^{m_k\times K_k}$ are the feature embedding matrices, with $\sum_{k=1}^{|F|}{m_k}=m$, and $K_k$ is the vocabulary size of the $k$th feature.
In other words, we look up separate embedding vectors for each feature, which are then concatenated.
The length of the concatenated vector matches the total embedding size, and all other parts of the model remain unchanged.

\section{Linguistic Input Features}

Our generalized model of the previous section supports an arbitrary number of input features.
In this paper, we will focus on a number of well-known linguistic features.
Our main empirical question is if providing linguistic features to the encoder improves the translation quality of neural machine translation systems, or if the information emerges from training encoder-decoder models on raw text, making its inclusion via explicit features redundant.
All linguistic features are predicted automatically;
we use Stanford CoreNLP \cite{N03-1033,DBLP:journals/nle/MinnenCP01,chen-manning:2014:EMNLP2014} to annotate the English input for English$\to$German, and ParZu \cite{sennrich13c} to annotate the German input for German$\to$English.
We here discuss the individual features in more detail.

\subsection{Lemma}

Using lemmas as input features guarantees sharing of information between word forms that share the same base form.
In principle, neural models can learn that inflectional variants are semantically related, and represent them as similar points in the continuous vector space \cite{conf/naacl/MikolovYZ13}.
However, while this has been demonstrated for high-frequency words, 
we expect that a lemmatized representation increases data efficiency; low-frequency variants may even be unknown to word-level models.
With character- or subword-level models, it is unclear to what extent they can learn the similarity between low-frequency word forms that share a lemma, especially if the word forms are superficially dissimilar.
Consider the following two German word forms, which share the lemma \textit{liegen} `lie':

\begin{itemize}
\item \textit{liegt} `lies' (3.p.sg.\ present)
\item \textit{läge} `lay' (3.p.sg. subjunctive II)
\end{itemize}

The lemmatisers we use are based on finite-state methods, which ensures a large coverage, even for infrequent word forms.
We use the Zmorge analyzer for German \cite{schmid2004,sennrich14},
and the lemmatiser in the Stanford CoreNLP toolkit for English \cite{DBLP:journals/nle/MinnenCP01}.

\subsection{Subword Tags}
\label{subword-sec}

In our experiments, we operate on the level of subwords to achieve open-vocabulary translation with a fixed symbol vocabulary,
using a segmentation based on \emph{byte-pair encoding} (BPE) \cite{DBLP:journals/corr/SennrichHB15}.
We note that in BPE segmentation, some symbols are potentially ambiguous, and can either be a separate word, or a subword segment of a larger word.
Also, text is represented as a sequence of subword units with no explicit word boundaries, but word boundaries are potentially helpful to learn which symbols to attend to, and when to forget information in the recurrent layers.
We propose an annotation of subword structure similar to popular IOB format for chunking and named entity recognition, marking if a symbol in the text forms the beginning (B), inside (I), or end (E) of a word.
A separate tag (O) is used if a symbol corresponds to the full word.

\subsection{Morphological Features}

For German$\to$English, the parser annotates the German input with morphological features.
Different word types have different sets of features -- for instance, nouns have case, number and gender, while verbs have person, number, tense and aspect -- and features may be underspecified.
We treat the concatenation of all morphological features of a word, using a special symbol for underspecified features, as a string, and treat each such string as a separate feature value.

\subsection{POS Tags and Dependency Labels}

In our introductory examples, we motivated POS tags and dependency labels as possible disambiguators.
Each word is associated with one POS tag, and one dependency label.
The latter is the label of the edge connecting a word to its syntactic head, or 'ROOT' if the word has no syntactic head.

\subsection{On Using Word-level Features in a Subword Model}

\begin{figure*}
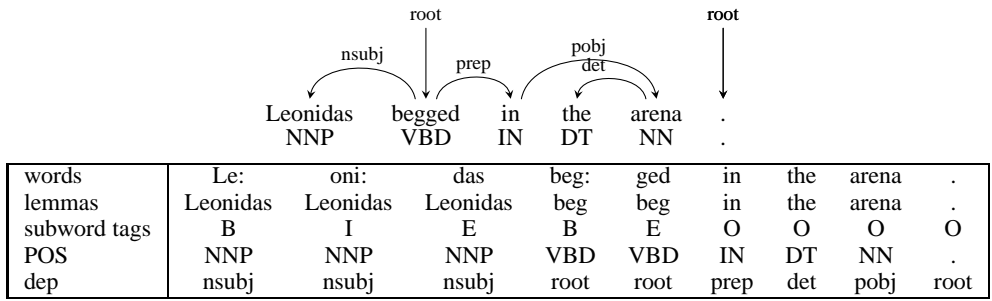

\centering
\small
\begin{dependency}[theme = simple, label style={scale=1.5}]
   \begin{deptext}[column sep=10pt]
      Leonidas \& begged \& in \& the \& arena \& . \\
      NNP \& VBD \& IN \& DT \& NN \& .\\
   \end{deptext}
   \deproot[edge unit distance=2ex]{2}{root}
   \deproot[edge unit distance=2ex]{6}{root}
   \depedge{2}{1}{nsubj}
   \depedge{2}{3}{prep}
   \depedge[label style={anchor=south east}]{5}{4}{det}
   \depedge{3}{5}{pobj}
   \deproot[edge unit distance=2ex]{6}{root}
\end{dependency}
\begin{tabular}{|l|ccccccccc|}
\hline
words & Le: & oni: & das & beg: & ged & in & the & arena & .\\
lemmas & Leonidas & Leonidas & Leonidas & beg & beg & in & the & arena & .\\
subword tags & B & I & E & B & E & O & O & O & O \\
POS & NNP & NNP & NNP & VBD & VBD & IN & DT & NN & .\\
dep & nsubj & nsubj & nsubj & root & root & prep & det & pobj & root \\
\hline
\end{tabular}
\caption{Original dependency tree for sentence \emph{Leonidas begged in the arena .}, and our feature representation after BPE segmentation.}
\label{example-figure}
\end{figure*}

We segment rare words into subword units using BPE.
The subword tags encode the segmentation of words into subword units, and need no further modification.
All other features are originally word-level features.
To annotate the segmented source text with features, we copy the word's feature value to all its subword units.
An example is shown in Figure \ref{example-figure}.

\section{Evaluation}
\label{sec:eval}

We evaluate our systems on the WMT16 shared translation task English$\leftrightarrow$German.
The parallel training data consists of about 4.2 million sentence pairs.

To enable open-vocabulary translation, we encode words via joint BPE\footnote{\url{https://github.com/rsennrich/subword-nmt}} \cite{DBLP:journals/corr/SennrichHB15}, learning \num{89500} merge operations on the concatenation of the source and target side of the parallel training data.
We use minibatches of size 80, a maximum sentence length of 50, word embeddings of size 500, and hidden layers of size 1024.
We clip the gradient norm to 1.0 \cite{DBLP:conf/icml/PascanuMB13}.
We train the models with Adadelta \cite{DBLP:journals/corr/abs-1212-5701}, reshuffling the training corpus between epochs.
We validate the model every \num{10000} minibatches via {\sc Bleu} and perplexity on a validation set (newstest2013).

For neural MT, perplexity is a useful measure of how well the model can predict a reference translation given the source sentence.
Perplexity is thus a good indicator of whether input features provide any benefit to the models, and we report the best validation set perplexity of each experiment.
To evaluate whether the features also increase translation performance,
we report case-sensitive {\sc Bleu} scores with mteval-13b.perl on two test sets, newstest2015 and newstest2016.
We also report {\sc chrF3} \cite{popovic:2015:WMT}, a character n-gram F$_3$ score which was found to correlate well with human judgments, especially for translations out of English 
\cite{stanojevic-EtAl:2015:WMT}.\footnote{We use the re-implementation included with the subword code}
The two metrics may occasionally disagree, partly because they are highly sensitive to the length of the output.
{\sc Bleu} is precision-based, whereas {\sc chrF3} considers both precision and recall, with a bias for recall.
For {\sc Bleu}, we also report whether differences between systems are statistically significant according to a bootstrap resampling significance test \cite{riezler-maxwell:2005:MTSumm}.

We train models for about a week, and report results for an ensemble of the 4 last saved models (with models saved every 12 hours).
The ensemble serves to smooth the variance between single models.

Decoding is performed with beam search with a beam size of 12.

\begin{table}
\centering
\small
\setlength\tabcolsep{3pt}
\begin{tabular}{l|rrr|rr}
& \multicolumn{3}{c|}{input vocabulary} & \multicolumn{2}{c}{embedding} \\
feature & EN & DE & model & all & single \\
\hline
subword tags & 4 & 4 & 4 & 5 & 5\\
POS tags & 46 & 54 & 54 & 10 & 10\\
morph.\ features & - & 1400 & 1400 & 10 & 10\\
dependency labels & 46 & 33 & 46& 10 & 10\\
lemmas & 800000 & 1500000 & 85000 & 115 & 167\\
words & 78500 & 85000 & 85000 & * & *\\
\end{tabular}
\caption{Vocabulary size, and size of embedding layer of linguistic features, in system that includes all features, and contrastive experiments that add a single feature over the baseline. The embedding layer size of the word feature is set to bring the total size to 500.}
\label{dim}
\end{table}

To ensure that performance improvements are not simply due to an increase in the number of model parameters, we keep the total size of the embedding layer fixed to 500.
Table \ref{dim} lists the embedding size we use for linguistic features -- the embedding layer size of the word-level feature varies, and is set to bring the total embedding layer size to 500.
If we include the lemma feature, we roughly split the embedding vector one-to-two between the lemma feature and the word feature.
The table also shows the network vocabulary size; for all features except lemmas, we can represent all feature values in the network vocabulary -- in the case of words, this is due to BPE segmentation.
For lemmas, we choose the same vocabulary size as for words, replacing rare lemmas with a special UNK symbol.

\newcite{2015arXiv151106709S} report large gains from using monolingual in-domain training data, automatically back-translated into the source language to produce a synthetic parallel training corpus.
We use the synthetic corpora produced in these experiments\footnote{The corpora are available at \url{http://statmt.org/rsennrich/wmt16_backtranslations/}} (3.6--4.2 million sentence pairs), and we trained systems which include this data to compare against the state of the art.
We note that our experiments with this data entail a syntactic annotation of automatically translated data, which may be a source of noise.
For the systems with synthetic data, we double the training time to two weeks.

We also evaluate linguistic features for the lower-resourced translation direction English$\to$Romanian, with 0.6 million sentence pairs of parallel training data, and 2.2 million sentence pairs of synthetic parallel data.
We use the same linguistic features as for English$\to$German.
We follow \newcite{sennrich-wmt16} in the configuration, and use dropout for the English$\to$Romanian systems.
We drop out full words (both on the source and target side) with a probability of 0.1.
For all other layers, the dropout probability is set to 0.2.

\subsection{Results}

\begin{table*}
\centering
\small
\begin{tabular}{l|c|cc|cc||c|cc|cc}
\multirow{3}{*}{system} & \multicolumn{5}{c||}{German$\to$English} & \multicolumn{5}{c}{English$\to$German}\\
& ppl $\downarrow$ & \multicolumn{2}{c|}{{\sc Bleu} $\uparrow$} & \multicolumn{2}{c||}{{\sc chrF3} $\uparrow$} & ppl $\downarrow$ & \multicolumn{2}{c|}{{\sc Bleu} $\uparrow$} & \multicolumn{2}{c}{{\sc chrF3} $\uparrow$}\\
& dev & test15 & test16  & test15 & test16 & dev & test15 & test16  & test15 & test16\\
\hline
baseline & 47.3 & 27.9\phantom{*} & 31.4\phantom{*} & 54.0 & 58.0 & 
54.9 & 23.0\phantom{*} & 27.8\phantom{*} & 52.6 & 56.0\\ 
all features & 46.2 & 28.7* & 32.9* & 54.8 & 58.5 & 
52.9 & 23.8* & 28.4* & 53.9 & 57.2\\ 
\end{tabular}
\caption{German$\leftrightarrow$English translation results: best perplexity on dev (newstest2013), and {\sc Bleu} and {\sc chrF3} on test15 (newstest2015) and test16 (newstest2016). {\sc Bleu} scores that are significantly different (p < 0.05) from respective baseline are marked with (*).}
\label{results-parallel}
\end{table*}

Table \ref{results-parallel} shows our main results for German$\to$English, and English$\to$German.
The baseline system is a neural MT system with only one input feature, the (sub)words themselves.
For both translation directions, linguistic features improve the best perplexity on the development data (47.3 $\rightarrow$ 46.2, and 54.9 $\rightarrow$ 52.9, respectively).
For German$\to$English, the linguistic features lead to an increase of 1.5 {\sc Bleu} (31.4$\rightarrow$32.9) and 0.5 {\sc chrF3} (58.0 $\rightarrow$ 58.5), on the newstest2016 test set.
For English$\to$German, we observe improvements of 0.6 {\sc Bleu} (27.8 $\rightarrow$ 28.4) and 1.2 {\sc chrF3} (56.0 $\rightarrow$ 57.2).

\begin{table*}
\centering
\small
\begin{tabular}{l|c|cc|cc||c|cc|cc}
\multirow{3}{*}{system} & \multicolumn{5}{c||}{German$\to$English} & \multicolumn{5}{c}{English$\to$German}\\
& ppl $\downarrow$ & \multicolumn{2}{c|}{{\sc Bleu} $\uparrow$} & \multicolumn{2}{c||}{{\sc chrF3} $\uparrow$} & ppl $\downarrow$ & \multicolumn{2}{c|}{{\sc Bleu} $\uparrow$} & \multicolumn{2}{c}{{\sc chrF3} $\uparrow$}\\
& dev & test15 & test16  & test15 & test16 & dev & test15 & test16  & test15 & test16\\
\hline
baseline & 47.3 & 27.9\phantom{*} & 31.4\phantom{*} & 54.0 & 58.0 & 
54.9 & 23.0\phantom{*} & 27.8\phantom{*} & 52.6 & 56.0 \\ 
lemmas & 47.1 & 28.4\phantom{*} & 32.3* & 54.6 & 58.7 & 
53.4 & 23.8* & 28.5* & 53.7 & 56.7 \\ 
subword tags & 47.3 & 27.7\phantom{*} & 31.5\phantom{*} & 54.0 & 58.1 & 
54.7 & 23.6* & 28.1\phantom{*} & 53.2 & 56.4\\ 
morph. features & 47.1 & 28.2\phantom{*} & 32.4* & 54.3 & 58.4 & 
- & - & - & - & - \\
POS tags & 46.9 & 28.1\phantom{*} & 32.4* & 54.1 & 57.8 & 
53.2 & 24.0* & 28.9*  & 53.3 & 56.8 \\ 
dependency labels & 46.9 & 28.1\phantom{*} & 31.8* & 54.2 & 58.3 & 
54.0 & 23.4*  & 28.0\phantom{*} & 53.1 & 56.5 \\ 
\end{tabular}
\caption{Contrastive experiments with individual linguistic features: best perplexity on dev (newstest2013), and {\sc Bleu} and {\sc chrF3} on test15 (newstest2015) and test16 (newstest2016). {\sc Bleu} scores that are significantly different (p < 0.05) from respective baseline are marked with (*).}
\label{contrastive-results}
\end{table*}

To evaluate the effectiveness of different linguistic features in isolation, we performed contrastive experiments in which only a single feature was added to the baseline.
Results are shown in Table \ref{contrastive-results}.
Unsurprisingly, the combination of all features (Table \ref{results-parallel}) gives the highest improvement, averaged over metrics and test sets, but most features are beneficial on their own.
Subword tags give small improvements for English$\to$German, but not for German$\to$English.
All other features outperform the baseline in terms of perplexity, and yield significant improvements in {\sc Bleu} on at least one test set.
The gain from different features is not fully cumulative; we note that the information encoded in different features overlaps.
For instance, both the dependency labels and the morphological features encode the distinction between German subjects and accusative objects, the former through different labels (\emph{subj} and \emph{obja}), the latter through grammatical case (\emph{nominative} and \emph{accusative}).

\begin{table*}
\centering
\small
\begin{tabular}{l|c|cc|cc||c|cc|cc}
\multirow{3}{*}{system} & \multicolumn{5}{c||}{German$\to$English} & \multicolumn{5}{c}{English$\to$German}\\
& ppl $\downarrow$ & \multicolumn{2}{c|}{{\sc Bleu} $\uparrow$} & \multicolumn{2}{c||}{{\sc chrF3} $\uparrow$} & ppl $\downarrow$ & \multicolumn{2}{c|}{{\sc Bleu} $\uparrow$} & \multicolumn{2}{c}{{\sc chrF3} $\uparrow$}\\
& dev & test15 & test16  & test15 & test16 & dev & test15 & test16  & test15 & test16\\
\hline
PBSMT \cite{williams2016} & - & 29.9\phantom{*} & 35.1\phantom{*} & 56.2 & 60.9 
& - & 23.7\phantom{*} & 28.4\phantom{*} & 52.6 & 56.6 \\ 
SBSMT \cite{williams2016} & - & 29.5\phantom{*} & 34.4\phantom{*} & 56.0 & 61.0 
& - & 24.5\phantom{*} & 30.6\phantom{*} & 55.3 & 59.9 \\ 
\hline
baseline & 45.2 & 31.5\phantom{*} & 37.5\phantom{*} & 57.0 & 62.2 & 
49.7 & 27.5\phantom{*} & 33.1\phantom{*} & 56.3 & 60.5\\ 
all features & 44.1 & 32.1* & 38.5* & 57.5 & 62.8 & 
48.4 & 27.1\phantom{*} & 33.2\phantom{*} & 56.5 & 60.6 \\ 
\end{tabular}
\caption{German$\leftrightarrow$English translation results with additional, synthetic training data: best perplexity on dev (newstest2013), and {\sc Bleu} and {\sc chrF3} on test15 (newstest2015) and test16 (newstest2016). {\sc Bleu} scores that are significantly different (p < 0.05) from respective baseline are marked with (*).}
\label{results-synthetic}
\end{table*}

We also evaluated adding linguistic features to a stronger baseline, which includes synthetic parallel training data.
In addition, we compare our neural systems against phrase-based (PBSMT) and syntax-based (SBSMT) systems by \cite{williams2016}, all of which make use of linguistic annotation on the source and/or target side.
Results are shown in Table \ref{results-synthetic}.
For German$\to$English, we observe similar improvements in the best development perplexity (45.2 $\rightarrow$ 44.1), test set {\sc Bleu} (37.5$\rightarrow$38.5) and {\sc chrF3} (62.2 $\rightarrow$ 62.8).
Our test set {\sc Bleu} is on par to the best submitted system to this year's WMT 16 shared translation task, which is similar to our baseline MT system, but which also uses a right-to-left decoder for reranking \cite{sennrich-wmt16}.
We expect that linguistic input features and bidirectional decoding are orthogonal, and that we could obtain further improvements by combining the two.

For English$\to$German, improvements in development set perplexity carry over (49.7 $\rightarrow$ 48.4),
but we see only small, non-significant differences in {\sc Bleu} and {\sc chrF3}.
While we cannot clearly account for the discrepancy between perplexity and translation metrics,
factors that potentially lower the usefulness of linguistic features in this setting are the stronger baseline,
trained on more data, and the low robustness of linguistic tools in the annotation of the noisy, synthetic data sets.
Both our baseline neural MT systems and the systems with linguistic features substantially outperform phrase-based and syntax-based systems for both translation directions.

\begin{figure}

\begin{tikzpicture}[scale=0.9]
\pgfplotsset{major grid style={style=dotted,color=black!20}}
\begin{axis}[xlabel=training time (minibatches $\cdot 10000$),
    xmin = 0,
    xmax = 60,
    ylabel=perplexity,
    legend pos = north east,
    legend style={
        font=\scriptsize,
        /tikz/nodes={anchor=west}
        },
    mark size = 0.1,
    ]

    \addplot +[black, no markers, raw gnuplot, solid, line width=0.2ex, id=baseline-ende] gnuplot {plot 'data/en-de1.plot';};
    \addplot +[black, no markers, raw gnuplot, dotted, line cap=round, line width=0.2ex, id=all-ende] gnuplot {plot 'data/en-de12.plot';};

    \addplot +[red, no markers, raw gnuplot, solid, line width=0.2ex, id=baseline-deen] gnuplot {plot 'data/de-en1.plot';};
    \addplot +[red, no markers, raw gnuplot, dotted, line cap=round, line width=0.2ex, id=all-deen] gnuplot {plot 'data/de-en5.plot';};

    \addlegendentry{EN-DE baseline (synth.\ data)}
    \addlegendentry{EN-DE all features (synth.\ data)}

    \addlegendentry{DE-EN baseline (synth.\ data)}
    \addlegendentry{DE-EN all features (synth.\ data)}

\end{axis}
\end{tikzpicture} 
\caption{English$\to$German (black) and German$\to$English (red) development set perplexity as a function of training time (number of minibatches) with and without linguistic features.}
\label{perpl}
\end{figure}
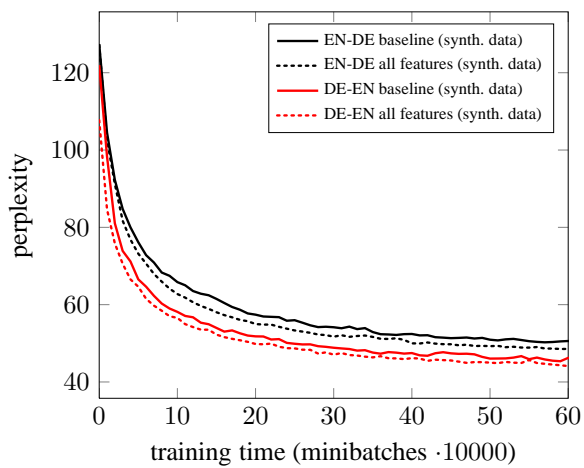

In the previous tables, we have reported the best perplexity.
To address the question about the randomness in perplexity, and whether the best perplexity just happened to be lower for the systems with linguistic features,
we show perplexity on our development set as a function of training time for different systems (Figure \ref{perpl}).
We can see that perplexity is consistently lower for the systems trained with linguistic features.

\begin{table}
\centering
\small
\begin{tabular}{l|c|c|c}
system & ppl $\downarrow$ & {\sc Bleu} $\uparrow$ & {\sc chrF3} $\uparrow$\\
\hline
\cite{qt21syscomb2016} & - & 28.9\phantom{*} & 57.1 \\
\hline
baseline & 74.9 & 23.8\phantom{*} & 52.5 \\ 
all features & 72.7 & 24.8* & 53.5 \\ 
\hline
baseline (+synth.\ data) & 50.9 & 28.2\phantom{*} & 56.1 \\ 
all features (+synth.\ data) & 50.1 & 29.2* & 56.6  \\ 
\end{tabular}
\caption{English$\to$Romanian translation results: best perplexity on newsdev2016, and {\sc Bleu} and {\sc chrF3} on newstest2016. {\sc Bleu} scores that are significantly different (p < 0.05) from respective baseline are marked with (*).}
\label{results-enro}
\end{table}

Table \ref{results-enro} shows results for a lower-resourced language pair, English$\to$Romanian.
With linguistic features, we observe improvements of 1.0 {\sc Bleu} over the baseline, both for the systems trained on parallel data only (23.8$\rightarrow$24.8),
and the systems which use synthetic training data (28.2$\rightarrow$29.2).
According to {\sc Bleu}, the best submission to WMT16 was a system combination by \newcite{qt21syscomb2016}.
Our best system is competitive with this submission.

Table \ref{examples} shows translation examples of our baseline, and the system augmented with linguistic features.
We see that the augmented neural MT systems, in contrast to the respective baselines, successfully resolve the reordering for the German$\to$English example,
and the disambiguation of \emph{close} for the English$\to$German example.

\begin{table}
\centering
\small
\setlength\tabcolsep{3pt}
\begin{tabular}{l|l}
system & sentence\\
\hline
source & Gefährlich ist die \textbf{Route} aber dennoch. \\ 
reference & However \textbf{the route} is dangerous.\\
baseline & Dangerous is \textbf{the route}, however.\\
all features & However, \textbf{the route} is dangerous.\\
\hline
\hline
source & [We thought] a win like this might be \textbf{close}.\\ 
reference & [...] dass ein solcher Gewinn \textbf{nah} sein könnte.\\
baseline & [...] ein Sieg wie dieser könnte \textbf{schließen}.\\
all features & [...] ein Sieg wie dieser könnte \textbf{nah} sein. \\
\hline
\end{tabular}
\caption{Translation examples illustrating the effect of adding linguistic input features.}
\label{examples}
\end{table}

\section{Related Work}

Linguistic features have been used in neural language modelling \cite{alexandrescu-kirchhoff:2006:HLT-NAACL06-Short}, and are also used in other tasks for which neural models have recently been employed, such as syntactic parsing \cite{chen-manning:2014:EMNLP2014}.
This paper addresses the question whether linguistic features on the source side are beneficial for neural machine translation.
On the target side, linguistic features are harder to obtain for a generation task such as machine translation, since this would require incremental parsing of the hypotheses at test time, and this is possible future work.

Among others, our model incorporates information from a dependency annotation, but is still a sequence-to-sequence model.
\newcite{2016arXiv160306075E} propose a tree-to-sequence model whose encoder computes vector representations for each phrase in the source tree.
Their focus is on exploiting the (unlabelled) structure of a syntactic annotation, whereas we are focused on the disambiguation power of the functional dependency labels.

Factored translation models are often used in  phrase-based SMT \cite{koehn-hoang:07} as a means to incorporate extra linguistic information. However, neural MT can provide a much more flexible
mechanism for adding such information.
Because phrase-based models cannot easily generalize to new feature combinations, the individual models either treat each feature combination as an atomic unit, resulting in data sparsity,
or assume independence between features, for instance by having separate language models for words and POS tags.
In contrast, we exploit the strong generalization ability of neural networks, and expect that even new feature combinations, e.g.\ a word that appears in a novel syntactic function, are handled gracefully.

One could consider the lemmatized representation of the input as a second source text, and perform multi-source translation \cite{2016arXiv160100710Z}.
The main technical difference is that in our approach, the encoder and attention layers are shared between features, which we deem appropriate for the types of features that we tested.

\section{Conclusion}

In this paper we investigate whether
linguistic input features are beneficial to neural machine translation, and our empirical evidence suggests that this is the case.

We describe a generalization of the encoder in the popular attentional encoder-decoder architecture for neural machine translation that allows for the inclusion of an arbitrary number of input features. We empirically test the inclusion of various linguistic features, including lemmas, part-of-speech tags, syntactic dependency labels, and morphological features, into English$\leftrightarrow$German, and English$\to$Romanian neural MT systems.
Our experiments show that the linguistic features yield improvements over our baseline, resulting in improvements on newstest2016 of 1.5 {\sc Bleu} for German$\to$English, 0.6 {\sc Bleu} for English$\to$German, and 1.0 {\sc Bleu} for English$\to$Romanian.

In the future, we expect several developments that will shed more light on the usefulness of linguistic (or other) input features, and whether they will establish themselves as a core component of neural machine translation.
On the one hand, the machine learning capability of neural architectures is likely to increase, decreasing the benefit provided by the features we tested.
On the other hand, there is potential to explore the inclusion of novel features for neural MT, which might prove to be even more helpful than the ones we investigated,
and the features we investigated may prove especially helpful for some translation settings, such as very low-resourced settings and/or translation settings with a highly inflected source language.

\section*{Acknowledgments}

This project has received funding from the European Union's Horizon 2020 research and innovation
programme under grant agreements 645452 (QT21), and 644402 (HimL).

\bibliographystyle{acl2016}
\bibliography{bibliography}

\end{document}